\relax
\documentclass[letterpaper]{article} 
\usepackage{aaai21}  
\usepackage{times}  
\usepackage{helvet} 
\usepackage{courier}  
\usepackage[hyphens]{url}  
\usepackage{graphicx} 
\usepackage{natbib}  
\usepackage{caption} 
\usepackage{amsmath}
\usepackage{makecell}
\usepackage{booktabs}
\usepackage{arydshln}
\usepackage{amsfonts}
\usepackage[switch]{lineno}
\nocopyright
\title{Knowledgeable Dialogue Reading Comprehension on Key Turns}

\author {
        Junlong Li\textsuperscript{\rm 1,2,3},
        Zhuosheng Zhang\textsuperscript{\rm 1,2,3},
        Hai Zhao\textsuperscript{\rm 1,2,3,\footnote{ Corresponding author.  This paper was partially supported by National Key Research and Development Program
of China (No. 2017YFB0304100), Key Projects of National Natural Science Foundation of China (U1836222 and
61733011), Huawei-SJTU long term AI project, Cutting-edge
Machine reading comprehension and language model.}}
        \\
}
\affiliations {
    \textsuperscript{\rm 1} Department of Computer Science and Engineering, Shanghai Jiao Tong University \\
    \textsuperscript{\rm 2} Key Laboratory of Shanghai Education Commission for Intelligent Interaction \\and Cognitive Engineering, Shanghai Jiao Tong University, Shanghai, China \\
    \textsuperscript{\rm 3} MoE Key Lab of Artificial Intelligence, AI Institute, Shanghai Jiao Tong University, Shanghai, China \\
    {\tt \{lockonn,zhangzs\}@sjtu.edu.cn, zhaohai@cs.sjtu.edu.cn}
}

\begin{document}
	\maketitle
	\begin{abstract}
		Multi-choice machine reading comprehension (MRC) requires a model to choose the correct answer from a set of answer options given passage and a question. When the given passage is derived from multi-turn dialogue, the task becomes even more challenging for multi-turn dialogue is extremely noisy and understanding dialogue casts especial prerequisite of commonsense knowledge which is unseen in the given material of the task. This work thus makes the first attempt to tackle those two challenges by extracting substantially important turns and utilizing external knowledge to enhance the representation of context. In our model, the relevance of each turn to the question is measured, and turns with high relevance are chosen as key turns. Besides, terms related to the context and the question in a knowledge graph are extracted as external knowledge. The key turns and the external knowledge are combined together with a well-designed mechanism to predict the answer. Experimental results on a dialogue-based MRC task, DREAM, show that our proposed model achieves great improvements on baselines.
	\end{abstract}
	\section{Introduction}
	
	\label{intro}

	Multi-choice Machine Reading Comprehension (MRC) is a challenging problem that has aroused great research interest. For the task, various datasets and models have been proposed \cite{lai2017race,ostermann2018semeval,khashabi2018looking,mostafazadeh2016corpus,richardson2013mctest,sun2019dream}. Formally, the task requires a model to select the best answer from a set of answer options given a passage and a question. 
	
	In conventional multi-choice MRC, passages are usually short articles   \cite{zhang2019dcmn,jin2019mmm,zhu2020dual,shoeybi2020megatron}. Recently, there come multi-choice MRC tasks whose passages are from multi-turn dialogues. Understanding multi-turn dialogues is much more complex than understanding short articles for two key challenges. Multi-turn dialogues are multi-party, multi-topic, and always have lots of turns in real-world cases (\textit{e.g.} chat history in social media). Besides, people rarely state obvious commonsense explicitly in their dialogues \cite{forbes2017verb}, therefore only considering superficial contexts may ignore implicit meaning and cause misunderstanding.
	
	\begin{table}
		\centering
		{
			\begin{tabular}{p{7.2cm}}
				\hline 
				\textbf{Dialogue 1}\\
				\hline
				\textit{W: \textbf{Well, I'm afraid my cooking isn't to your}}  \textit{\textbf{taste.}}                \\
				\textit{M: \textbf{Actually, I like it very much.}}                         \\
				\textit{W: \underline{I'm glad you say that. Let me serve you some }}
				\textit{\underline{more fish.}}                          \\
				
				\textit{M: \textbf{Thanks. I didn't know you are so good at cooking.}} \\ 
				
				\textit{W: Why not bring your wife next time?}                                    \\
				\textit{M: OK, I will. She will be very glad to see } \textit{you, too.}                                \\
				\hline
				Question: \textit{What does the man think of the }
				\textit{woman's cooking?}                            \\
				\hline
				\textit{A. It's really terrible.}                                                   \\
				\textit{B. It's very good indeed. *}                                                                  \\
				\textit{C. It's better than what he does.}                                                            \\
				\hline
			\end{tabular}%
		}
		\makeatletter\def\@captype{table}\makeatother\caption{Turns in different format have different topics and those in bold are key turns.}
		\label{tb-example-topic}
	\end{table}

	We demonstrate two examples from DREAM dataset \cite{sun2019dream} in Tables \ref{tb-example-topic} and \ref{tb:example-commonsense}. Table \ref{tb-example-topic} shows the topic shifts in turns and only a few of them, called key turns, are directly related to the question. Treating all turns equally brings no benefit to understanding of multi-turn dialogues as shown in some previous works \cite{zhang2018modeling,yuan2019multi}. Table \ref{tb:example-commonsense} shows that commonsense knowledge are required to answer the question (\textit{e.g.} bikes are always on the streets). However, such required knowledge usually cannot be obtained or inferred from the given material (passage, question or answer options) of the task.
	
	Prior work in DREAM \cite{sun2019dream} first investigated the effects of incorporating dialogue structure and different kinds of general commonsense knowledge into both rule-based and machine learning-based reading comprehension models, acquiring significant improvement compared to its baseline models. Recently, more studies have achieved remarkable progress on dialogue-based multi-choice MRC tasks \cite{zhu2020dual,jin2019mmm,wang2019evidence}. Generally, all these existing models consist of two components: encoder and matching network. A critical feature of these models is that they choose advanced pre-trained language models (PrLMs) such as BERT \cite{devlin2019bert} and GPT \cite{radford2018improving} as encoder implementation.     
	
	\begin{table}
		\centering
		{
			\begin{tabular}{p{7cm}}
				\hline
				\textbf{Dialogue 2}\\
				\hline
				\textit{\textbf{M: Look at the girl on the bike!}}\\
				\textit{F: Oh, yes she's really a smart girl.}\\
				\hline
				{Question: \textit{Where are the two persons?}}                            \\
				\hline
				\textit{A. At home.}                                                   \\
				\textit{B. In their classroom. }                                                                  \\
				\textit{C. On the street. *}                                                            \\
				\hline
			\end{tabular}
		}
		\makeatletter\def\@captype{table}\makeatother\caption{Commonsense is required in the question.}
		\label{tb:example-commonsense}
	\end{table}
	
	To face the most challenges posed by the latest dialogue reading comprehension \cite{sun2019dream}, we propose modeling with \textbf{K}nowledge and \textbf{K}ey \textbf{T}urns (KKT) to solve this type of MRC problem better. In detail, our model utilizes the key turns concerned the question and external commonsense knowledge like knowledge graphs to enhance the language representation based on the PrLM encoder. It further predicts answers through a well-designed matching network. A PrLM fine-tuned for natural language inference (NLI) is used to extract key turns from the context by following successful practice in previous work \cite{phang2018sentence,conneau2017supervised,subramanian2018learning} so that the resulted PrLM can figure out how large the relevance is between a turn and a specific question. Besides, the needed knowledge is extracted from a knowledge graph to enrich the context like K-BERT \cite{liu2019k} and ERNIE \cite{zhang2019ernie}. Then the extracted key turns and knowledge are used to refine the language representations of the context, question, and candidate answers for our matching network.
	
	The contributions of this paper are summarized as follows:
	
	1. We adopt an explainable and efficient method to extract key turns from multi-turn dialogues.
	
	2. We introduce external knowledge from a knowledge graph to enrich its representation. 
	
	3. Experimental results on the DREAM dataset show our model can achieve significant improvement over the strong baseline.
	
	The remainder of the paper is organized as follows. Section \ref{sec:rw} reviews related work. In Section \ref{sec:method} we describe details of KKT inspired model. Sections \ref{sec:exp} and \ref{sec:ana} show the experiments and analysis respectively, and the conclusion of this study is in Section \ref{sec:conc}.
	
	\section{Related Work}\label{sec:rw}
	\subsection{Pre-trained Language Model}
	\begin{figure*}
		\centering
		\includegraphics[scale=0.8]{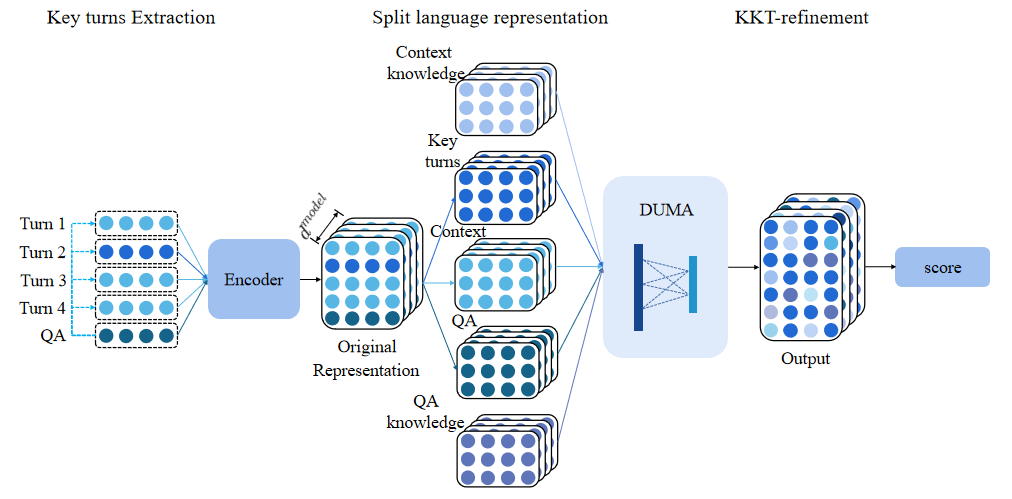}
		\caption{Overall model architecture}
		\label {overallmodel}
	\end{figure*}
	Recently, pre-trained Language Models have been shown to be effective in learning universal language representations, achieving state-of-the-art results in a series of flagship
	natural language processing (NLP) tasks. Prominent examples are Embedding from Language Models (ELMo) \cite{peters2018deep}, Generative Pre-trained Transformer (OpenAI GPT) \cite{radford2018improving}, BERT \cite{devlin2019bert}, Generalized Autoregressive Pre-training (XLNet) \cite{yang2019xlnet}, Roberta \cite{liu2019roberta} and ALBERT \cite{lan2019albert}. These models follow a pre-training and fine-tuning uasge: first pre-trained on large unlabeled corpus with specific training targets (\textit{e.g.} Masked LM and Next Sentence Prediction for BERT), then fine-tuned on different downstream tasks with reletively small labeled corpus. 
	
	\subsection{Multi-choice Machine Reading Comprehension}
	
	A lot of multi-choice Machine Reading Comprehension (MRC) datasets have been proposed \cite{richardson2013mctest,mostafazadeh2016corpus,khashabi2018looking,ostermann2018semeval,lai2017race,sun2019dream}. The majority of questions in multi-choice MRC tasks are non-extractive, requiring multi-sentence summarizing, reasoning and additional commonsense knowledge. Correspondingly, many models tackling multi-choice MRC were proposed. \citet{jin2019mmm} chose a PrLM as the encoder and used a multi-step attention network as the classifier on top of the encoder. \citet{sun2019dream} designed specific neural models for dialogue-based multi-choice MRC by adding a speaker embedding to enhance the representation of context on word-level. \citet{zhang2019dcmn} and \citet{zhu2020dual} took a powerful PrLM as the encoder and used a bidirectional matching network to predict the answer.

	\subsection{Knowledge Enhanced Language Representation}
	
	Recently, researchers pay more and more attention to enhancing text models with Knowledge Graphs (KGs), since KGs obtain a great amount of systematic knowledge.  Integrating background knowledge in a neural model was first proposed in the neural-checklist model by \citeauthor{kiddon2016globally} \shortcite{kiddon2016globally} for text generation of recipes. \cite{liu2019k} combined knowledge triples in KGs with original texts before modeling them with BERT to get more hidden information. \citeauthor{mihaylov2018knowledgeable}\shortcite{mihaylov2018knowledgeable} attended to relevant external knowledge and combined this knowledge with the context representation in Cloze-style reading comprehension. \citeauthor{bosselut2019comet}\shortcite{bosselut2019comet} employed the triples in KGs as corpus to train GPT \cite{radford2018improving} for commonsense learning. \citeauthor{lin2019kagnet}\shortcite{lin2019kagnet} proposed a knowledge-aware graph network based on GCN and LSTM with a path-based attention mechanism. \citeauthor{zhang2019ernie}\shortcite{zhang2019ernie} fused entity information with BERT to enhance language representation, which can take advantage of lexical, syntactic, and knowledge information simultaneously.
	
	Some previous works have already taken either key turns \cite{yuan2019multi} or commonsense knowledge \cite{sun2019dream} into account when dealing with multi-turn dialogues. However, none of them make a combination of these two important factors. In this work, we make the first attempt to utilize key turns and commonsense knowledge simultaneously to enhance the language representation of multi-turn dialogues. Related knowledge items in knowledge graphs are rewritten and encoded with the PrLM. On the other hand, representations of key turns are directly extracted from the encoded text of the original dialogues. 
	As a result, representations of additional commonsense knowledge, key turns, and the original text of multi-turn dialogues are in the same vector space so that they can be easily fused together.

	\section{Methodology}\label{sec:method}
	
	The overall architecture of our model is shown in Figure \ref{overallmodel}.

	\subsection{Notations}
	The context $C$ for the concerned multi-turn dialogue MRC may be represented as $[T_1,T_2...T_{l_c}]$. For each turn $T_i$ in the context, it may be represented as $[w_1,w_2...w_{t_i}]$, where $t_i$ is the number of tokens in $T_i$. The question $Q$ is represented as $[w_1,w_2...w_q]$ where $q$ is the number of tokens in $Q$. The answer options $A$ are represented as $[A_1,A_2..A_{l_a}]$ and each one of them is represented as $[w_1,w_2,...w_{a_j}]$ where $a_i$ is the number of tokens in $j$-th answer option $A_j$. In this work, we treat the question and the answer option as an integral, so that we have $QA_j=[Q: A_j]$. Therefore the task aim is to find the most proper question-answer pair according to the context. 
	
	Multi-head attention \cite{vaswani2017attention} is used in this work to capture the relationship between two sequences. We denote it as MHA(·), which is implemented as follows:
	\begin{equation}
	\begin{aligned}
	&\textrm{Att}(E'_Q,E'_K,E'_V)=\textrm{softmax}(\frac{E'_Q(E'_K)^T}{\sqrt{d_{head}}})E'^V,\\
	&\textrm{head$_i$}=\textrm{Att}(E_QW^Q_i,E_KW^K_i,E_VW^V_i),\\
	&\textrm{MHA}(E_Q,E_K,E_V)=\textrm{Concat(head$_i$...head$_h$)},\\
	\end{aligned}
	\end{equation}
	where $W^Q_i \in \mathbb{R}^{d_{model} \times d_{ head}}, W^K_i \in \mathbb{R}^{d_{model} \times d_{head}}, W^V_i \in \mathbb{R}^{d_{model} \times d_{head}}, E_Q \in \mathbb{R}^{d_q \times d_{model}}, E_K \in \mathbb{R}^{d_k \times d_{model}}, E_V \in \mathbb{R}^{d_v \times d_{model}}$, $d_q,d_k,d_v$ and $d_{head}$ denote the dimension of Query vectors, Key vectors, Value vectors and each head, respectively. $h$ denotes the number of heads, and we always assume $d_k=d_v$ and $d_{model}=h \times d_{head}$.

	\subsection{Extracting Knowledge}
	Items with weight less than a threshold or contain words that are not in the vocabulary of the chosen PrLM are removed from KG. The items are triples with the form \{\textit{relation, head, tail}\}, which are rewritten as facts (\textit{e.g. \{causes, virus, disease\}} to \textit{virus causes disease}). These facts are encoded with our adopted PrLM and the last hidden states $H_k $ ($H_k\in \mathbb{R}^{n\times d_{model}}$ where \textit{n} denotes the number of tokens in the fact) are taken as the output so that the representations of knowledge and context are in the same vector space. A self-attention module is used to refine the representation of each fact. We use mean-pooling in the end to aggregate the representation of each token and get a final representation $r_k$ ($r_k \in \mathbb{R}^{d_{model}}$) for each fact.
	
	\begin{equation}
	\begin{aligned}
	&\textrm{SelfAttention}(H_k)=\textrm{MHA}(H_k,H_k,H_k),\\
	&r_k=\textrm{mean}(\textrm{SelfAttention}(H_k)).
	\end{aligned}
	\end{equation}
	
	\subsection{Retrieve Relevant Knowledge}
	Each turn $T_i$ is tagged part-of-speech (POS) tags.  For tokens with POS like adjective, noun, and verb, we assume that they contain more implicit information than others; thus, items related to them are retrieved in KG. In all the chosen items, top $p$ (a hyperparameter) ones are selected to enhance the context representation. The extracted knowledge items are denoted as $CK=[r_{c_1},...r_{c_p}]$.
	
	For a QA-pair $QA_j$, we follow the same steps in dealing with $T_i$ to get the relevant knowledge items $QAK_j=[r_{j_1},r_{j_2},...r_{j_k}]$, where $j_k$ is the number of chosen knowledge items for $QA_j$.
	
	\subsection{Extracting Key Turns}
	Key turns are extracted with a PrLM fine-tuned for NLI. Each turn $T_i$ and each QA-pair $QA_j$ are concatenated and encoded with the NLI-fine-tuned LM. Pooled output are mapped into 3 dimensions, corresponding to \textit{contradiction}, \textit{entailment}, and \textit{neutral} respectively. Dimension of \textit{entailment} is chosen as the relevance score between $T_i$ and $QA_j$, and top $k$ turns $T_j^{key}$ are picked out as key turns for $QA_j$ according to the relevance scores ($k$ denotes the maximum number of key turns for each QA-pair). 
	
	\subsection{Encoding and Representation Refinement}
	For each $QA_j$ , it is concatenated with $C$ as input encoded with PrLM. The last hidden states $H_t \in \mathbb{R}^{l_{input}\times d_{model}}$ are then separated into context representation $H_c \in \mathbb{R}^{l_c\times d_{model}}$ and QA-pair representation $H_{QA}\in \mathbb{R}^{l_{QA}\times d_{model}}$. 
	
	The representation of key turns $H_{kt}$ ($H_{kt} \in \mathbb{R}^{l_{kt} \times d_{model}}$) is extracted from $H_c$ based on the position of key turns in the context. We use \textit{MHA(·)} to calculate the key-turns-refined context representation and  $H_{c_{kt}}$ ($H_{c_{kt}} \in \mathbb{R}^{l_c \times d_{model}}$). Simlilarly, we get the knowledge-refined representation of context and QA-pair:
	
	\begin{equation}
	\begin{aligned}
	&H_{c_{kt}}=\textrm{MHA}(H_c,H_{kt},H_{kt}),\\
	&H_{c_{k}}=\textrm{MHA}(H_c,CK,CK),\\
	&H_{QA_{k}}=\textrm{MHA}(H_{QA},QAK,QAK).
	\end{aligned}
	\end{equation}

	\subsection{Representation Fusion}
	Following \citet{zhu2020dual}, we use a Dual Multi-head Co-Attention (DUMA) module to fuse the representation of context and QA-pair.
	\begin{equation}
	\begin{aligned}
	&\textrm{MHA$_1$}=\textrm{MHA}(H_c,H_{QA},H_{QA}),\\
	&\textrm{MHA$_2$}=\textrm{MHA}(H_{QA},H_{QA},H_c),\\
	&\textrm{DUMA}(H_c, H_{QA}) = \textrm{Concat(mean(MHA$_1$)}\\&\textrm{,mean(MHA$_2$))}.
	\end{aligned}
	\end{equation}
	
	Based on different represenation of context and QA-pair calculated above, DUMA module may give three types of outputs, the original $O_O$, the key-turns refined $O_{kt}$ and the knowledge refined $O_k$ as follow.
	\begin{equation}
	\begin{aligned}
	&O_o=\textrm{DUMA}(H_c,H_{QA}),\\
	&O_{kt}=\textrm{DUMA}(H_{c_{kt}},H_{QA}),\\
	&O_k=\textrm{DUMA}(H_{c_{k}},H_{QA_k}).
	\end{aligned}
	\end{equation}
	
	Then these three kind of outputs are fused together as the final output. $O_k$ and $O_{kt}$ are concatenated together and mapped to dimension of $2d_{model}$ through a linear layer to get the knowledge-key-turns refined (KKT-refined) output $O_{kkt}$. Then we fuse the original output and the KKT-refined output to get the final output $O$. Concatenation is chosen as our fuse function.
	
	\subsection{Decoding}
	Our model decoder takes $O$ and computes the probability distribution over answer options. Let $A_i$ be the $i$-th answer option and $O_i$ is the corresponding output of $<C,Q,A_i>$. The loss function is computed by
	
	\begin{equation}
	L(Ai|C,Q)=-\textrm{log}(\frac{\textrm{exp}(W^TO_i)}{\sum_{j=1}^{l_a}\textrm{exp}(W^TO_j)}),
	\end{equation}
	where $W$ is a learnable parameter.
	
	\section{Experiments}\label{sec:exp}
	\subsection{Dataset}
	We evaluate our model on the DREAM \cite{sun2019dream} dataset, which is the newly released dialogue-based multi-choice MRC dataset. The detailed data statistics are listed in Table \ref{tb:data-stat}. 
	
	\begin{table}
		\centering
		\setlength{\tabcolsep}{15pt}
		{
			\begin{tabular}{ll}
				\hline
				& \textbf{DREAM} \\
				\hline
				\# of dialogues            & 6444           \\
				\# of questions            & 10197          \\
				Extractive(\%)             & 16.3           \\
				Abstractive(\%)            & 83.7           \\
				Average answer length      & 5.3            \\
				\# of answers per question & 
				3\\      
				\hline       
			\end{tabular}
		}
		\makeatletter\def\@captype{table}\makeatother\caption{Data statistics of the DREAM dataset.}
		\label{tb:data-stat}
	\end{table}
	
	DREAM is collected from English exams. Each dialogue, as the given context has multiple questions, and each question has three answer options. The most important feature of the dataset is that most of the questions are non-extractive. As a result, the dataset is small but quite challenging.

	\subsection{Settings}
	
	Our model uses a powerful PrLM ALBERT$_{xxlarge}$ \cite{lan2019albert} as our encoder, which is implemented based on Transformers\footnote{\url{https://github.com/huggingface/transformers}.}. Results of the ALBERT model as baseline are our rerunning unless otherwise specified. In this paper, NLTK \cite{loper2002nltk} is adopted as our POS tagger and ConceptNet \cite{speer2017conceptnet} is chosen as our KG. The NLI dataset used to fine-tune the PrLM for key turns extracting is SNLI \cite{bowman2015large}, and the accuracy of this fine-tuned model on the test set of SNLI is 90\%. 
	
	The evaluation metric we use is accuracy, \emph{acc=N$^+$/N}, where \textit{N$^+$} denotes the number of examples the model selects the correct answer, and \textit{N} denotes the total number of evaluation examples. 
	
	The learning rate in our model is 1e-5, training batch size per GPU is 1, and warmup steps are 50. We train the model for 3 epochs on 8 NVIDIA V100 GPUs. In the following Section \ref{sec:ana}, for the models used for further analysis based on ALBERT$_{base}$, our learning rate is 1e-5, with no warmup steps, and the training batch size per-GPU is 2. For all the above experiments, checkpoints are saved after each epoch, and the best results on the dev set are chosen from all the checkpoints for test evaluation.
	
	\begin{table}
		
		\centering
		\setlength{\tabcolsep}{10pt}
		{
			\begin{tabular}{lll}
				\hline
				\textbf{Model}                                                  & \textbf{Dev} & \textbf{Test} \\
				\hline
				FTLM++                                       & 58.1$^\star$     & 58.2$^\star$      \\
				BERT$_{large}$                             & 66.0$^\star$     & 66.8$^\star$      \\
				XLNet                                       & -         & 72.0$^\star$      \\
				RoBERTa$_{large}$                          & 85.4$^\star$     & 85.0$^\star$      \\
				RoBERTa$_{large}$+MMM                        & 88.0$^\star$     & 88.9$^\star$      \\
				ALBERT$_{xxlarge}$                        & 89.2$^\star$     & 88.5$^*$      \\
				ALBERT$_{xxlarge}$+DUMA                       & 89.3$^\dagger$     & \textbf{90.4}$^\dagger$      \\
				\hline
				ALBERT$_{xxlarge}$                               & 89.1          & 88.2  \\
				ALBERT$_{xxlarge}$+KKT                                   & \textbf{90.2}         & 89.8  \\
				ALBERT$_{base}$                                   & 67.4         & 67.3 \\
				ALBERT$_{base}$+KKT                                       & 69.3        & 68.7 \\
				\hline        
			\end{tabular}%
		}
		\caption{\label{font-table} Results on DREAM dataset. Results denoted by $\star$ are from \cite{jin2019mmm}, $\dagger$ are from \cite{zhu2020dual}. 
			\label{tb:main}
		}
	\end{table}
	
	\subsection{Main Results}
	
	Table \ref{tb:main} gives the main results of our experiments. To focus on the strength of models themselves, we only compare with single models instead of ensemble and multi-tasking ones. Experimental results show our model obtains a great improvement compared to the baseline and achieves state-of-the-art performance for DREAM on dev set. 
	
	\section{Analysis}\label{sec:ana}
	\subsection{Effects of External Knowledge Appending}
	
	Multi-turn dialogues more or less involve with lots of implicit or explicit commonsense between lines, therefore understanding them requires the support of knowledge. In our model, a knowledge graph is used by adding related knowledge items. In the example in Table  \ref{tb:example-commonsense} (in section \ref{intro}), we need to know where a bike is likely to appear for answering the question. Correspondingly, a knowledge item \textit{\{atlocation, bike, street\}} can be found in our KG, which means \textit{Bikes are always found on the street}. It has a weight of 2, indicating it is more important than others with a lower weight. With such a fact, the model thus can answer the question correctly.  
	
\begin{figure}
		\centering
		\includegraphics[scale=0.37]{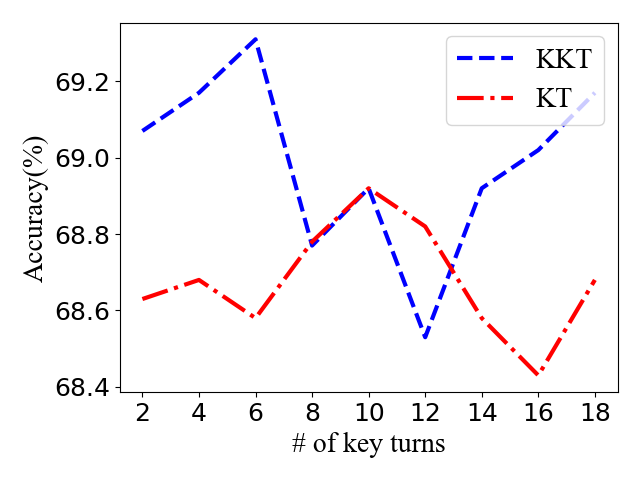}
		\caption{Comparison between complete model (KKT) and model without knowledge-refinement (KT) on different numbers of key turns}
		\label{fig:com-kt}
	\end{figure}

	To state the effects more clearly, we remove the knowledge refined output from our model and evaluate it on the dev set. The results are shown in Figure \ref{fig:com-kt}. We can see a general performance improvement from the knowledge refined part for different numbers of key turns.
	
	\subsection{Effects of Key Turns Extraction}
	
	As mentioned in the previous section, a challenge in understanding and modeling multi-turn dialogues is the topic shift in different turns, which means only a few turns are truly related to the question. By using an NLI-fine-tuned PrLM, we can get the relevance score for each turn-QA pair. A higher relevance score indicates that it is more likely to conclude the QA given the corresponding turn. The example shown in Table \ref{tb:score} proves our hypothesis. Turns with top 2 entailment scores can directly give the answer, while the other turns have nothing to do with the answer. This example shows that key turns are decisive for context refinement in representation.
	
	\begin{table}
		\centering
		\begin{tabular}{ll}
			\hline 
			\textbf{Dialogue 1}                                                         & \textbf{Score} \\
			\hline
			\textit{W: Well, I'm afraid my cooking} \\ 
			\textit{isn't to your taste.}                & -1.91                 \\
			\textit{\textbf{M: Actually, I like it very much.}}                         & \textbf{-1.49}                     \\
			\textit{W: I'm glad you say that. Let me}\\
			\textit{serve you some more fish.}         & -2.53                     \\
			
			\textit{\textbf{M: Thanks. I didn't know you}}\\ 
			\textit{\textbf{were so good at cooking.}}     & \textbf{-1.66}                     \\
			\textit{W: Why not bring your wife next time?}                              & -2.26                     \\
			\textit{M: OK, I will. She will be very glad}\\
			\textit{to see you, too.}              & -1.87                     \\
			\hline
			Question: \textit{What does the man think}\\
			\textit{of the woman's cooking?} &                           \\
			\hline
			\textit{A. It's really terrible.}                                                   \\
			\textit{B. It's very good indeed. *}                                        &                           \\
			\textit{C. It's better than what he does.}                                  &                          \\
			\hline
		\end{tabular}%
		\caption{Relevance score for each turn corresponding to the correct answer.}
		\label{tb:score}
	\end{table}
	
	To address the effects more clearly, we evaluate our model by removing the key-turns refined output from our model and evaluate it. The maximum number of knowledge items is 30. The results are shown in Table \ref{tb:complete}, which indicates significant performance loss on both dev and test sets.
	
	\begin{table}
		\centering
		\setlength{\tabcolsep}{13pt}
		{
			\begin{tabular}{lll}
				\hline
				\textbf{Model}                     & \textbf{Dev} & \textbf{Test} \\
				\hline
				ALBERT$_{base}$                    & 67.40         & 67.31         \\
				ALBERT$_{base}$+KKT                    & 69.32         & 68.71         \\
				\qquad \qquad \qquad \ -KT & \textbf{67.94}        & \textbf{67.66}         \\
				\hline
			\end{tabular}%
		}
		\caption{\label{tb:complete}Comparison between complete model, model without key-turns-refinement and baseline.}
	\end{table}

	Note that our preprocessing on PrLM benefits from another NLP task, NLI, which shows different down-stream NLP tasks are closely related. In our work, the NLI task serves as an indispensable part in our modeling, indicating the possibility that we can obtain a better language representation by cascading different models and taking advantage of each one of them. 
	
	\subsection{Effects of Number of Key Turns}
	
\begin{figure}
		\centering
		\includegraphics[scale=0.37]{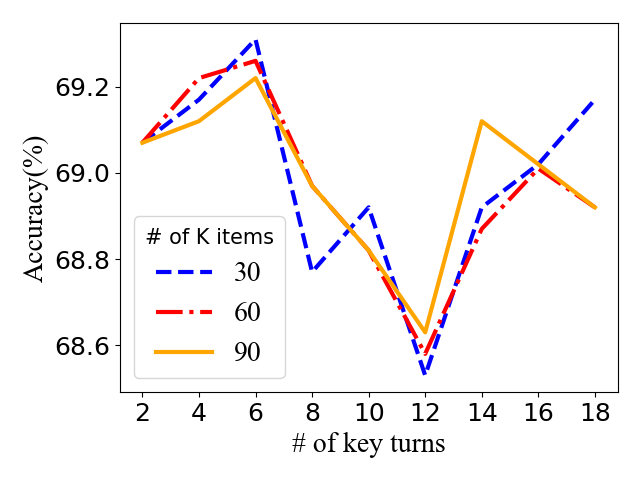}
		\caption{Different number of knowledge items (K items) and different numbers of key turns.}
		\label{fig:kkt}
	\end{figure}

	To find out the relationship between the performance of our model and the number of key turns, we evaluate our model on different numbers of extracted turns on dev set. The results on dev set are shown in Figure \ref{fig:kkt}. 
	
	The results in Figure \ref{fig:kkt} show that no matter how many key turns are chosen, obvious improvement can be observed in all cases. The performance peak occurs when the number of key turns is 6, which means when the key turns are unnecessarily too many for accurate enough representation. In that case, key turns can help filter the noise in other turns to a great extent. When more key turns are chosen, the effects of filtering noise are weaker, and the computational cost rises, so a small number of key turns in our model is preferred to refine context.

	\subsection{Effects of Number of Knowledge Items}
	The number of knowledge Items can affect performance as well. So we evaluate our model on different numbers of knowledge items. The results are also shown in Figure \ref{fig:kkt}. Contrary to our expectation, the results show little difference in various numbers of knowledge items. We suppose that the attention mechanism employed to refine context with external knowledge would contribute to the result because knowledge items with small weight tend to be concerned less in the attention mechanism as well since they are less relevant to the context. Therefore, the knowledge-refined context is similar though using more numbers of knowledge items, leading to a similar final performance when choosing different number of knwowledge items.   
	
	\subsection{Only Encoding Key Turns}
\begin{figure}
		\centering
		\includegraphics[scale=0.37]{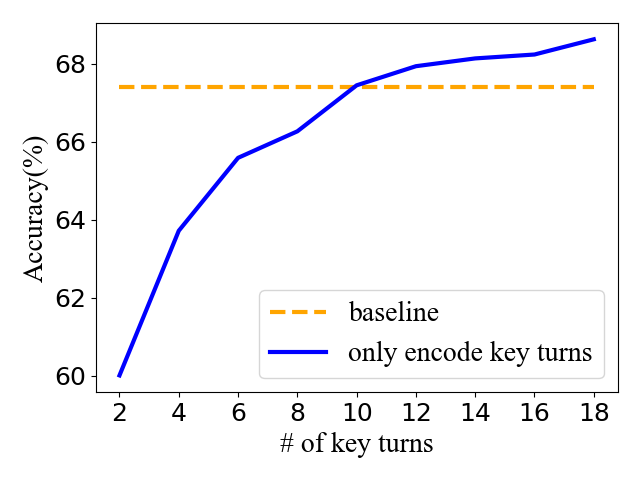}
		\caption{Only encoding key turns and omitting unrelated turns.}
		\label{fig:kt}
	\end{figure}

	For seeking the best performance, it is natural to figure out only encoding the selected key turns and omitting all the unrelated turns, which will reduce the computational cost a lot with much fewer tokens encoded. So we only choose key turns as our context and evaluate our model on dev set. The results are shown in Figure \ref{fig:kt}. 
	
	The comparison in Figure \ref{fig:kt} shows that the performance is much worse than the baseline when the number of key turns is too small. But when it goes beyond 10, the performance approaches and even surpasses the baseline. It indicates that by designing methods properly, we can indeed use only some key turns of the dialogue for better performance, as well as reducing input sequence length and computation cost.
	
	\section{Conclusion}\label{sec:conc}
	In this paper, we propose modeling multi-turn dialogue with knowledge and key turns for dialogue-based multi-choice MRC. We first pick out some of the turns from the dialogue directly related to the question as key turns. Then, the relevant knowledge items are picked out and encoded with PrLM. Question and context are refined with key turns and external knowledge items for better language representation. Experiments on benchmark multi-turn dialogue MRC dataset, DREAM show our model achieves great improvements compared to the baseline model. The procedures of our method are highly explainable and reflect a primary idea of cascading different models to get better language representation. Experiments also show that omitting unrelated turns and only encoding key turns has certain effectiveness and reduces computational cost a lot.

	\bibliography{reference.bib}

\begin{thebibliography}{33}
\providecommand{\natexlab}[1]{#1}
\providecommand{\url}[1]{\texttt{#1}}
\providecommand{\urlprefix}{URL }
\expandafter\ifx\csname urlstyle\endcsname\relax
  \providecommand{\doi}[1]{doi:\discretionary{}{}{}#1}\else
  \providecommand{\doi}{doi:\discretionary{}{}{}\begingroup
  \urlstyle{rm}\Url}\fi

\bibitem[{Bosselut et~al.(2019)Bosselut, Rashkin, Sap, Malaviya, Celikyilmaz,
  and Choi}]{bosselut2019comet}
Bosselut, A.; Rashkin, H.; Sap, M.; Malaviya, C.; Celikyilmaz, A.; and Choi, Y.
  2019.
\newblock {COMET}: Commonsense Transformers for Automatic Knowledge Graph
  Construction.
\newblock In \emph{Proceedings of the 57th Annual Meeting of the Association
  for Computational Linguistics (ACL 2019)}, 4762--4779.
\newblock \doi{10.18653/v1/P19-1470}.
\newblock \urlprefix\url{https://www.aclweb.org/anthology/P19-1470}.

\bibitem[{Bowman et~al.(2015)Bowman, Angeli, Potts, and
  Manning}]{bowman2015large}
Bowman, S.; Angeli, G.; Potts, C.; and Manning, C.~D. 2015.
\newblock A large annotated corpus for learning natural language inference.
\newblock In \emph{Proceedings of the 2015 Conference on Empirical Methods in
  Natural Language Processing (EMNLP 2015)}, 632--642.
\newblock \doi{10.18653/v1/D15-1075}.
\newblock \urlprefix\url{https://www.aclweb.org/anthology/D15-1075}.

\bibitem[{Conneau et~al.(2017)Conneau, Kiela, Schwenk, Barrault, and
  Bordes}]{conneau2017supervised}
Conneau, A.; Kiela, D.; Schwenk, H.; Barrault, L.; and Bordes, A. 2017.
\newblock Supervised learning of universal sentence representations from
  natural language inference data.
\newblock In \emph{Proceedings of the 2017 Conference on Empirical Methods in
  Natural Language Processing (EMNLP 2017)}, 670--680.
\newblock \doi{10.18653/v1/D17-1070}.
\newblock \urlprefix\url{https://www.aclweb.org/anthology/D17-1070}.

\bibitem[{Devlin et~al.(2019)Devlin, Chang, Lee, and
  Toutanova}]{devlin2019bert}
Devlin, J.; Chang, M.-W.; Lee, K.; and Toutanova, K. 2019.
\newblock {BERT}: Pre-training of Deep Bidirectional Transformers for Language
  Understanding.
\newblock In \emph{Proceedings of the 2019 Conference of the North American
  Chapter of the Association for Computational Linguistics: Human Language
  Technologies, Volume 1 (Long and Short Papers) (NAACL 2019)}, 4171--4186.
\newblock \doi{10.18653/v1/N19-1423}.
\newblock \urlprefix\url{https://www.aclweb.org/anthology/N19-1423}.

\bibitem[{Forbes and Choi(2017)}]{forbes2017verb}
Forbes, M.; and Choi, Y. 2017.
\newblock Verb {P}hysics: Relative Physical Knowledge of Actions and Objects.
\newblock In \emph{Proceedings of the 55th Annual Meeting of the Association
  for Computational Linguistics (Volume 1: Long Papers) (ACL 2017)}, 266--276.
\newblock \doi{10.18653/v1/P17-1025}.
\newblock \urlprefix\url{https://www.aclweb.org/anthology/P17-1025}.

\bibitem[{Jin et~al.(2019)Jin, Gao, Kao, Chung, and Hakkani-tur}]{jin2019mmm}
Jin, D.; Gao, S.; Kao, J.-Y.; Chung, T.; and Hakkani-tur, D. 2019.
\newblock {MMM}: Multi-stage Multi-task Learning for Multi-choice Reading
  Comprehension.
\newblock \emph{arXiv preprint arXiv:1910.00458}
  \urlprefix\url{https://arxiv.org/abs/1910.00458}.

\bibitem[{Khashabi et~al.(2018)Khashabi, Chaturvedi, Roth, Upadhyay, and
  Roth}]{khashabi2018looking}
Khashabi, D.; Chaturvedi, S.; Roth, M.; Upadhyay, S.; and Roth, D. 2018.
\newblock Looking Beyond the Surface: A Challenge Set for Reading Comprehension
  over Multiple Sentences.
\newblock In \emph{Proceedings of the 2018 Conference of the North American
  Chapter of the Association for Computational Linguistics: Human Language
  Technologies, Volume 1 (Long Papers) (NAACL 2018)}, 252--262.
\newblock \doi{10.18653/v1/N18-1023}.
\newblock \urlprefix\url{https://www.aclweb.org/anthology/N18-1023}.

\bibitem[{Kiddon, Zettlemoyer, and Choi(2016)}]{kiddon2016globally}
Kiddon, C.; Zettlemoyer, L.; and Choi, Y. 2016.
\newblock Globally coherent text generation with neural checklist models.
\newblock In \emph{Proceedings of the 2016 Conference on Empirical Methods in
  Natural Language Processing (EMNLP 2016)}, 329--339.
\newblock \doi{10.18653/v1/D16-1032}.
\newblock \urlprefix\url{https://www.aclweb.org/anthology/D16-1032}.

\bibitem[{Lai et~al.(2017)Lai, Xie, Liu, Yang, and Hovy}]{lai2017race}
Lai, G.; Xie, Q.; Liu, H.; Yang, Y.; and Hovy, E. 2017.
\newblock {RACE}: Large-scale ReAding Comprehension Dataset From Examinations.
\newblock In \emph{Proceedings of the 2017 Conference on Empirical Methods in
  Natural Language Processing (EMNLP 2017)}, 785--794.
\newblock \doi{10.18653/v1/D17-1082}.
\newblock \urlprefix\url{https://www.aclweb.org/anthology/D17-1082}.

\bibitem[{Lan et~al.(2019)Lan, Chen, Goodman, Gimpel, Sharma, and
  Soricut}]{lan2019albert}
Lan, Z.; Chen, M.; Goodman, S.; Gimpel, K.; Sharma, P.; and Soricut, R. 2019.
\newblock {ALBERT}: A lite bert for self-supervised learning of language
  representations.
\newblock \emph{arXiv preprint arXiv:1909.11942}
  \urlprefix\url{https://arxiv.org/abs/1909.11942}.

\bibitem[{Lin et~al.(2019)Lin, Chen, Chen, and Ren}]{lin2019kagnet}
Lin, B.~Y.; Chen, X.; Chen, J.; and Ren, X. 2019.
\newblock {K}ag{N}et: Knowledge-Aware Graph Networks for Commonsense Reasoning.
\newblock In \emph{Proceedings of the 2019 Conference on Empirical Methods in
  Natural Language Processing and the 9th International Joint Conference on
  Natural Language Processing (EMNLP-IJCNLP 2019)}, 2822--2832.
\newblock \doi{10.18653/v1/D19-1282}.
\newblock \urlprefix\url{https://www.aclweb.org/anthology/D19-1282}.

\bibitem[{Liu et~al.(2019{\natexlab{a}})Liu, Zhou, Zhao, Wang, Ju, Deng, and
  Wang}]{liu2019k}
Liu, W.; Zhou, P.; Zhao, Z.; Wang, Z.; Ju, Q.; Deng, H.; and Wang, P.
  2019{\natexlab{a}}.
\newblock {K-BERT}: Enabling language representation with knowledge graph.
\newblock \emph{arXiv preprint arXiv:1909.07606}
  \urlprefix\url{https://arxiv.org/abs/1909.07606}.

\bibitem[{Liu et~al.(2019{\natexlab{b}})Liu, Ott, Goyal, Du, Joshi, Chen, Levy,
  Lewis, Zettlemoyer, and Stoyanov}]{liu2019roberta}
Liu, Y.; Ott, M.; Goyal, N.; Du, J.; Joshi, M.; Chen, D.; Levy, O.; Lewis, M.;
  Zettlemoyer, L.; and Stoyanov, V. 2019{\natexlab{b}}.
\newblock {RoBERTa}: A robustly optimized bert pretraining approach.
\newblock \emph{arXiv preprint arXiv:1907.11692}
  \urlprefix\url{https://arxiv.org/abs/1907.11692}.

\bibitem[{Loper and Bird(2002)}]{loper2002nltk}
Loper, E.; and Bird, S. 2002.
\newblock {NLTK}: The Natural Language Toolkit.
\newblock In \emph{Proceedings of the ACL-02 Workshop on Effective Tools and
  Methodologies for Teaching Natural Language Processing and Computational
  Linguistics}, 63--70.
\newblock \doi{10.3115/1118108.1118117}.
\newblock \urlprefix\url{https://www.aclweb.org/anthology/W02-0109}.

\bibitem[{Mihaylov and Frank(2018)}]{mihaylov2018knowledgeable}
Mihaylov, T.; and Frank, A. 2018.
\newblock Knowledgeable Reader: Enhancing Cloze-Style Reading Comprehension
  with External Commonsense Knowledge.
\newblock In \emph{Proceedings of the 56th Annual Meeting of the Association
  for Computational Linguistics (Volume 1: Long Papers) (ACL 2018)}, 821--832.
\newblock \doi{10.18653/v1/P18-1076}.
\newblock \urlprefix\url{https://www.aclweb.org/anthology/P18-1076}.

\bibitem[{Mostafazadeh et~al.(2016)Mostafazadeh, Chambers, He, Parikh, Batra,
  Vanderwende, Kohli, and Allen}]{mostafazadeh2016corpus}
Mostafazadeh, N.; Chambers, N.; He, X.; Parikh, D.; Batra, D.; Vanderwende, L.;
  Kohli, P.; and Allen, J. 2016.
\newblock A Corpus and Cloze Evaluation for Deeper Understanding of Commonsense
  Stories.
\newblock In \emph{Proceedings of the 2016 Conference of the North American
  Chapter of the Association for Computational Linguistics: Human Language
  Technologies (NAACL 2016)}, 839--849.
\newblock \doi{10.18653/v1/N16-1098}.
\newblock \urlprefix\url{https://www.aclweb.org/anthology/N16-1098}.

\bibitem[{Ostermann et~al.(2018)Ostermann, Roth, Modi, Thater, and
  Pinkal}]{ostermann2018semeval}
Ostermann, S.; Roth, M.; Modi, A.; Thater, S.; and Pinkal, M. 2018.
\newblock SemEval-2018 Task 11: Machine Comprehension Using Commonsense
  Knowledge.
\newblock In \emph{Proceedings of The 12th International Workshop on Semantic
  Evaluation (SemEval 2018)}, 747--757.
\newblock \doi{10.18653/v1/S18-1119}.
\newblock \urlprefix\url{https://www.aclweb.org/anthology/S18-1119}.

\bibitem[{Peters et~al.(2018)Peters, Neumann, Iyyer, Gardner, Clark, Lee, and
  Zettlemoyer}]{peters2018deep}
Peters, M.~E.; Neumann, M.; Iyyer, M.; Gardner, M.; Clark, C.; Lee, K.; and
  Zettlemoyer, L. 2018.
\newblock Deep contextualized word representations.
\newblock In \emph{Proceedings of the 2018 Conference of the North American
  Chapter of the Association for Computational Linguistics: Human Language
  Technologies, Volume 1 (Long Papers) (NAACL 2018)}, 2227--2237.
\newblock \doi{10.18653/v1/N18-1202}.
\newblock \urlprefix\url{https://www.aclweb.org/anthology/N18-1202}.

\bibitem[{Phang, F{\'e}vry, and Bowman(2018)}]{phang2018sentence}
Phang, J.; F{\'e}vry, T.; and Bowman, S.~R. 2018.
\newblock Sentence encoders on {STILTS}: Supplementary training on intermediate
  labeled-data tasks.
\newblock \emph{arXiv preprint arXiv:1811.01088}
  \urlprefix\url{https://arxiv.org/abs/1811.01088}.

\bibitem[{Radford et~al.(2018)Radford, Narasimhan, Salimans, and
  Sutskever}]{radford2018improving}
Radford, A.; Narasimhan, K.; Salimans, T.; and Sutskever, I. 2018.
\newblock Improving language understanding by generative pre-training.
\newblock \emph{Technical report, OpenAI}
  \urlprefix\url{https://www.semanticscholar.org/paper/Improving-Language-Understanding-by-Generative-Radford/cd18800a0fe0b668a1cc19f2ec95b5003d0a5035}.

\bibitem[{Richardson, Burges, and Renshaw(2013)}]{richardson2013mctest}
Richardson, M.; Burges, C.~J.; and Renshaw, E. 2013.
\newblock {MCT}est: A Challenge Dataset for the Open-Domain Machine
  Comprehension of Text.
\newblock In \emph{Proceedings of the 2013 Conference on Empirical Methods in
  Natural Language Processing (EMNLP 2013)}, 193--203.
\newblock \urlprefix\url{https://www.aclweb.org/anthology/D13-1020}.

\bibitem[{Shoeybi et~al.(2019)Shoeybi, Patwary, Ruri, LeGresley, Casper, and
  Catanzaro}]{shoeybi2020megatron}
Shoeybi, M.; Patwary, M.; Ruri, R.; LeGresley, P.; Casper, J.; and Catanzaro,
  B. 2019.
\newblock Megatron-{LM}: Training Multi-Billion Parameter Language Models Using
  Model Parallelism.
\newblock \emph{arXiv preprint arXiv:1909.08053}
  \urlprefix\url{https://arxiv.org/abs/1909.08053}.

\bibitem[{Speer, Chin, and Havasi(2017)}]{speer2017conceptnet}
Speer, R.; Chin, J.; and Havasi, C. 2017.
\newblock Concept{N}et 5.5: an open multilingual graph of general knowledge.
\newblock In \emph{Proceedings of the Thirty-First AAAI Conference on
  Artificial Intelligence (AAAI 2017)}, 4444--4451.
\newblock \urlprefix\url{https://arxiv.org/abs/1612.03975}.

\bibitem[{Subramanian et~al.(2018)Subramanian, Trischler, Bengio, and
  Pal}]{subramanian2018learning}
Subramanian, S.; Trischler, A.; Bengio, Y.; and Pal, C.~J. 2018.
\newblock Learning general purpose distributed sentence representations via
  large scale multi-task learning.
\newblock \emph{arXiv preprint arXiv:1804.00079}
  \urlprefix\url{https://arxiv.org/abs/1804.00079}.

\bibitem[{Sun et~al.(2019)Sun, Yu, Chen, Yu, Choi, and Cardie}]{sun2019dream}
Sun, K.; Yu, D.; Chen, J.; Yu, D.; Choi, Y.; and Cardie, C. 2019.
\newblock {DREAM}: A challenge data set and models for dialogue-based reading
  comprehension.
\newblock \emph{Transactions of the Association for Computational Linguistics}
  7: 217--231.
\newblock \doi{10.1162/tacl_a_00264}.
\newblock \urlprefix\url{https://www.aclweb.org/anthology/Q19-1014}.

\bibitem[{Vaswani et~al.(2017)Vaswani, Shazeer, Parmar, Uszkoreit, Jones,
  Gomez, Kaiser, and Polosukhin}]{vaswani2017attention}
Vaswani, A.; Shazeer, N.; Parmar, N.; Uszkoreit, J.; Jones, L.; Gomez, A.~N.;
  Kaiser, {\L}.; and Polosukhin, I. 2017.
\newblock Attention is all you need.
\newblock In \emph{Advances in neural information processing systems (NIPS
  2017)}, 5998--6008.
\newblock
  \urlprefix\url{https://papers.nips.cc/paper/7181-attention-is-all-you-need.pdf}.

\bibitem[{Wang et~al.(2019)Wang, Yu, Sun, Chen, Yu, McAllester, and
  Roth}]{wang2019evidence}
Wang, H.; Yu, D.; Sun, K.; Chen, J.; Yu, D.; McAllester, D.; and Roth, D. 2019.
\newblock Evidence Sentence Extraction for Machine Reading Comprehension.
\newblock In \emph{Proceedings of the 23rd Conference on Computational Natural
  Language Learning (CoNLL 2019)}, 696--707.
\newblock \doi{10.18653/v1/K19-1065}.
\newblock \urlprefix\url{https://www.aclweb.org/anthology/K19-1065}.

\bibitem[{Yang et~al.(2019)Yang, Dai, Yang, Carbonell, Salakhutdinov, and
  Le}]{yang2019xlnet}
Yang, Z.; Dai, Z.; Yang, Y.; Carbonell, J.; Salakhutdinov, R.~R.; and Le, Q.~V.
  2019.
\newblock {XLN}et: Generalized autoregressive pretraining for language
  understanding.
\newblock In \emph{Advances in neural information processing systems (NIPS
  2019)}, 5754--5764.
\newblock \urlprefix\url{https://arxiv.org/abs/1906.08237}.

\bibitem[{Yuan et~al.(2019)Yuan, Zhou, Li, Lv, Zhu, Han, and
  Hu}]{yuan2019multi}
Yuan, C.; Zhou, W.; Li, M.; Lv, S.; Zhu, F.; Han, J.; and Hu, S. 2019.
\newblock Multi-hop Selector Network for Multi-turn Response Selection in
  Retrieval-based Chatbots.
\newblock In \emph{Proceedings of the 2019 Conference on Empirical Methods in
  Natural Language Processing and the 9th International Joint Conference on
  Natural Language Processing (EMNLP-IJCNLP 2019)}, 111--120.
\newblock \doi{10.18653/v1/D19-1011}.
\newblock \urlprefix\url{https://www.aclweb.org/anthology/D19-1011}.

\bibitem[{Zhang et~al.(2019{\natexlab{a}})Zhang, Zhao, Wu, Zhang, Zhou, and
  Zhou}]{zhang2019dcmn}
Zhang, S.; Zhao, H.; Wu, Y.; Zhang, Z.; Zhou, X.; and Zhou, X.
  2019{\natexlab{a}}.
\newblock {DCMN}+: Dual co-matching network for multi-choice reading
  comprehension.
\newblock \emph{arXiv preprint arXiv:1907.11692}
  \urlprefix\url{https://arxiv.org/abs/1908.11511}.

\bibitem[{Zhang et~al.(2019{\natexlab{b}})Zhang, Han, Liu, Jiang, Sun, and
  Liu}]{zhang2019ernie}
Zhang, Z.; Han, X.; Liu, Z.; Jiang, X.; Sun, M.; and Liu, Q.
  2019{\natexlab{b}}.
\newblock {ERNIE}: Enhanced Language Representation with Informative Entities.
\newblock In \emph{Proceedings of the 57th Annual Meeting of the Association
  for Computational Linguistics (ACL 2019)}, 1441--1451.
\newblock \doi{10.18653/v1/P19-1139}.
\newblock \urlprefix\url{https://www.aclweb.org/anthology/P19-1139}.

\bibitem[{Zhang et~al.(2018)Zhang, Li, Zhu, Zhao, and Liu}]{zhang2018modeling}
Zhang, Z.; Li, J.; Zhu, P.; Zhao, H.; and Liu, G. 2018.
\newblock Modeling Multi-turn Conversation with Deep Utterance Aggregation.
\newblock In \emph{Proceedings of the 27th International Conference on
  Computational Linguistics (COLING 2018)}, 3740--3752.
\newblock \urlprefix\url{https://www.aclweb.org/anthology/C18-1317}.

\bibitem[{Zhu, Zhao, and Li(2020)}]{zhu2020dual}
Zhu, P.; Zhao, H.; and Li, X. 2020.
\newblock Dual Multi-head Co-attention for Multi-choice Reading Comprehension.
\newblock \emph{arXiv preprint arXiv:2001.09415}
  \urlprefix\url{https://arxiv.org/abs/2001.09415}.

\end{thebibliography}
\end{document}